# A Plausibility Study of Using Augmented Reality in the Ventriculoperitoneal Shunt Operations

Tandin Dorji[1], Pakinee Aimmanee[2]
School of Information, Computer, and Communication Technology,
Sirindhorn International Institute of Technology,
Thammasat University, Thailand.
m6622040068@g.siit.tu.ac.th[1], pakinee@siit.tu.ac.th[2]

Vich Yindeedej
Department of Neurosurgery,
Faculty of Medicine, Thammasat University,
Thammasat Chalermprakiat Hospital, Thailand
vichyindeedej@gmail.com

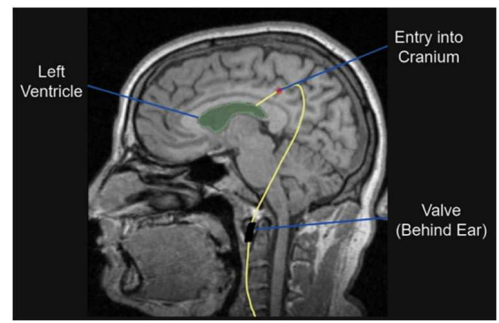

*Figure 1 VP Shunt model*

*Abstract*—The field of augmented reality (AR) has undergone substantial growth, finding diverse applications in the medical industry. This paper delves into various techniques employed in medical surgeries, scrutinizing factors such as cost, implementation, and accessibility. The focus of this exploration is on AR-based solutions, with a particular emphasis on addressing challenges and proposing an innovative solution for ventriculoperitoneal shunt (VP) operations. The proposed solution introduces a novel flow in the pre-surgery phase, aiming to substantially reduce setup time and operation duration by creating 3D models of the skull and ventricles. Experiments are conducted where the models are visualized on a 3D-printed skull through an AR device, specifically the Microsoft HoloLens 2. The paper then conducts an in-depth analysis of this proposed solution, discussing its feasibility, advantages, limitations, and future implications.

## I. Introduction

Hydrocephalus, a neurological disorder, arises from the accumulation of excess cerebrospinal fluid (CSF) within a patient's cerebral ventricles. This surplus CSF exerts pressure on the brain tissues, causing the ventricles to expand and leading to tissue damage and subsequent malfunctions [1]. With an annual incidence of up to 400,000 cases [2], hydrocephalus stems from various causes, including an imbalance in CSF production and absorption rates [3]. One prevalent treatment for hydrocephalus is ventriculoperitoneal shunting (VP shunting). This procedure involves drilling a small hole into the patient's head at a determined entry point, allowing the ventricular catheter (VC) to access the lateral ventricles and facilitate drainage of excess CSF to the abdominal cavity via the subcutaneous tract, as depicted in Figure 1.

In free-hand neurosurgical operations, identifying the precise entry point and trajectory holds paramount importance. The accuracy of these determinations is critical, as deviations can lead to unintended VC placement, potentially resulting in adverse effects such as damage to brain tissues and even a brain hemorrhage. Keen's Point, a common and easily locatable entry point, is typically measured 3 cm backward and 3 cm vertically upward from the top of the ear's helix. However, the inherent anatomical variations among individuals and the possibility of human error introduce challenges, leading to procedural inconsistencies and delays.

Advanced and expensive methods [4] employ tools that digitally scan the head, providing a 3D visualization of the brain to enhance accuracy in determining the entry point, trajectory, and shunt depth. Techniques such as stereotaxy, neuronavigation, ultrasound, electromagnetic navigation, and robotic navigation deliver precise results but are hindered by their high costs and space-intensive requirements, limiting accessibility, especially in resource-intensive settings.

An alternative approach involves computer-aided pre-surgery techniques that recommend an entry point without the overhead of intricate machinery. One notable example is the VP Shunt entry point recommender (VPSEAR) [4], representing a significant leap in computer-assisted systems for VP shunt operations. VPSEAR calculates the entry area precisely based on CT scan slices, offering a tailored and patient-specific approach. However, a limitation arises as the output is presented to the surgeon in a 2-

dimensional space, requiring them to divert attention during surgery by continually switching focus between the patient and a reference monitor elsewhere in the room.

This work addresses the challenge faced by surgeons by proposing a solution through Augmented Reality (AR) technology, extensively studied for brain operation purposes. The subsequent sections provide an insightful review of relevant studies in this domain, exploring the potential of AR to enhance the visualization of entry points and trajectories, ultimately improving precision and surgical outcomes.

AR has emerged as a transformative technology in the medical field, particularly in neurosurgery, offering potential enhancements in surgical precision and a streamlined learning curve. One notable study by Berger et al. explores the application of AR to address Trigeminal Neuralgia (TN) [5]. It aims to integrate AR seamlessly into medical practice for surgical interventions. The study highlights a significant success rate increase (90.6%) in cannulation procedures through AR, contrasting with the modest success rate (18.8%) observed with conventional techniques. The AR approach involves using a specialized headset to overlay computed tomography scans (CT scans), providing real-time guidance for needle insertion. This not only reduces surgical duration but also mitigates human error, marking a substantial advancement in procedural accuracy. However, the study prompts considerations about the cost-effectiveness of widespread AR adoption, emphasizing the need for further research into economic aspects.

In a related effort, Glas et al. address a critical challenge using 3D virtual surgical planning (3D VSP) [6], aiming to integrate it into the operating room environment through a real-time navigation system. The study compares the proposed 3D VSP approach with the industry standard, Brainlab, focusing on speed, accuracy, and usability. Utilizing Microsoft HoloLens for real-time visualization, the research introduces a paradigm shift in surgical practices, potentially reducing the workload on medical professionals and leading to improved patient outcomes. Responding to challenges in the external ventricular drain (EVD) procedure, Eom et al. present NeuroLens, an AR-based solution [7]. The EVD procedure's reliance on surgical skills often leads to varying success rates and complications, prompting the need for innovation. NeuroLens leverages real-time tracking, patient-specific ventricular holograms, and catheter placement guidance, catering to novice and experienced surgeons. Despite groundbreaking results, the study acknowledges limitations, emphasizing the ongoing commitment to refining NeuroLens as a transformative tool in neurosurgical interventions.

This work extends the efforts of VPSEAR, aiming to provide a clear visualization of the ventricles and, consequently, assist surgeons in locating the entry areas suggested by the VPSEAR program through Microsoft HoloLens 2. The implementation involves an AR Head-Mounted Display (HMD) attached to the surgeon's head, offering an augmented perspective and comprehensive contextual information beyond the scope of the unaided eye. This proposed solution holds particular promise for remote public hospitals worldwide that lack access to expensive medical equipment. By leveraging AR advancements, it aims to enhance surgical procedures and provide valuable support to surgeons, ultimately benefiting patient outcomes.

II. PROPOSED METHOD

The framework of our proposed approach is detected in Figure 2.

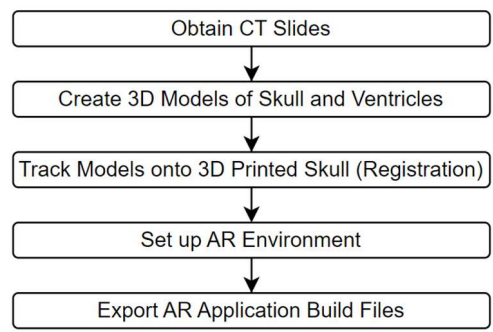

*Figure 2 Framework of creating the AR application with marker-based tracking technique*

The details of each step are described as follows. In this innovative surgical approach, the initial step involved obtaining the patient's CT slides, using DICOM files as the foundation for creating virtual models of the skull and ventricles through 3D slicing software. It is recommended to create two distinct models—one for the skull and another for the ventricles—enabling us to switch between visualizing the skull and ventricles with ease.

The next crucial step involved mapping these virtual representations onto the actual 3D-printed skull through a process called registration. While multiple tracking cameras could enhance accuracy, practical challenges in the limited operating room space make this option impractical. Similarly, employing object tracking with the 3D printed skull as a 3D model is deemed impractical due to the introduction of logistical complexities and an extended setup time for each operation. The chosen solution was marker-based tracking, where a marker was placed at a fixed distance from the skull, striking a balance between precision and practicality. This was the preferred option due to its time efficiency, cost-effectiveness, spatial constraints in the operating room, and considerations for patient well-being.

The 3D marker was integrated into a chosen 3D engine, forming the basis for developing the AR application. This application, constructed on a separate computer, uses the 3D engine to create models dynamically following the marker's position in the real world. To maintain a safe distance from the patient's head, the models were strategically positioned away from the 3D marker, reducing the risk of infection in the entry area.

Transferring build files from the computer to the AR device was typically seamless and did not impose a significant time burden on the overall preparation. We then equipped ourselves with the AR device compatible with the 3D engine and its build files to visualize the tracked models. Ensuring the 3D-printed skull was positioned at the same distance from the 3D marker as the models in the 3D engine allowed for an accurate overlay of the virtual skull and ventricles onto the 3D-printed skull, providing a clear visualization.

This streamlined process enabled easy visualization of the ventricles in the 3D-printed skull and simplified the visual confirmation of the alignment between the recommended entry point and the ventricles. If the recommended entry point aligns with the anatomical structure of the patient's head, the surgeon can promptly initiate the surgery. In cases of discrepancies, indicating potential anatomical variance, the surgeon can explore alternative approaches, minimizing the time needed to confirm algorithm compatibility.

### III. EXPERIMENTS AND PRELIMINARY RESULTS

The CT slides and the 3D-printed skull used in this experiment were supplied by Thammasat University Hospital. The data collection process involved the use of Philips IQon Elite Spectral and Philips Brilliance iCT 256-slice CT scanners. In the experiment, three different sets of CT slides were employed, with each set utilizing 243 CT slides. The skull and ventricles were modeled from these slides using a 3D Slicer [8], as illustrated in Figure 3.

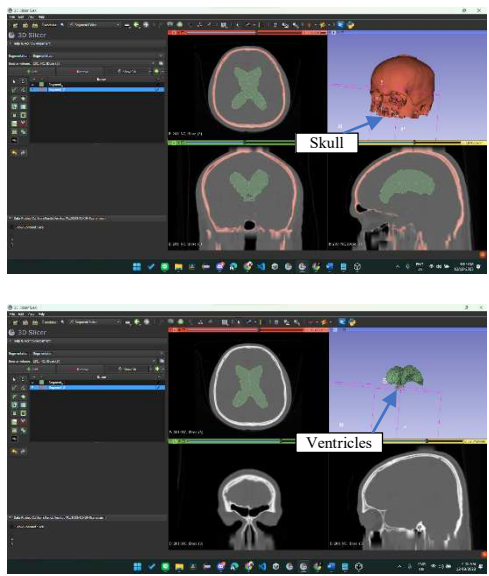

*Figure 3. Modelling skull (top) and ventricles (bottom) using 243 CT Slides in CT Slicer*

Vuforia [9] was strategically employed for the registration step because it had a user-friendly setup and straightforward integration with the selected 3D Engine, Unity [10], in conjunction with the mixed reality toolkit 3 (MRTK3) [11]. Within Vuforia, a range of tracking markers are provided and referred to as targets. However, to optimize accuracy for the current task, exploration experiments with image-based and cylindrical targets were conducted. Figure 4 illustrates the skull model after setting between Unity and Vuforia using the image target and the cylindrical target.

The AR device used in our experiment was Microsoft HoloLens 2 [12]. It is well adaptable with Unity, so it was perfect for constructing applications for the Windows Universal Platform. For developing an application for the HoloLens 2 the MRTK3 toolkit was chosen for its compatibility with HoloLens 2 and Unity.

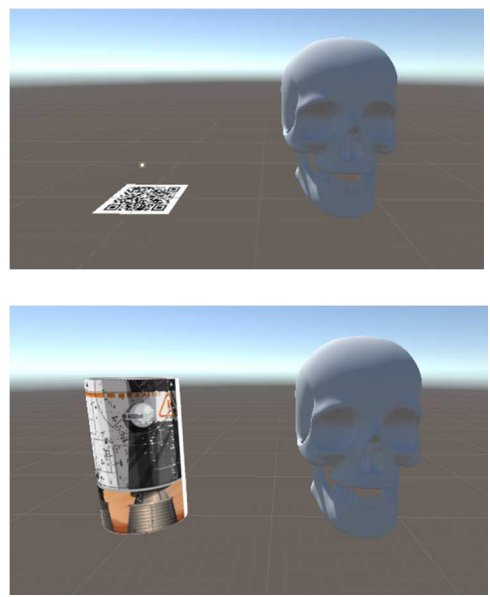

*Figure 4. Successful Unity and Vuforia setup to display the skull model using image target (top) and cylindrical target (bottom)*

A proof of concept (POC) was crafted to validate the proposed approach's feasibility. The POC entailed importing a simple skull and teeth model from the internet into Unity and tracking them using image and cylindrical targets, as depicted in Figure 5.

Two buttons were also incorporated into the POC, each toggling the display of the skull and teeth respectively. The integration of the MRTK3 toolkit also facilitated the easy incorporation of voice commands, allowing for toggling specific sections of the skull on and off using customizable keywords. The implemented voice commands, namely "Toggle Head" and "Toggle Teeth," were designed to activate the corresponding buttons.

The models were then positioned at a fixed distance of 10 cm from the targets and tested. Figure 6 shows a visual representation of the skull and marker placement and demonstrates the projection of the virtual skull onto the 3D-printed skull.

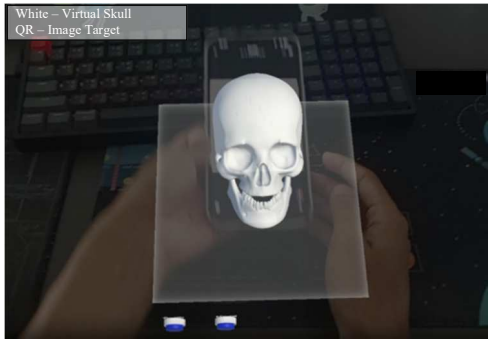

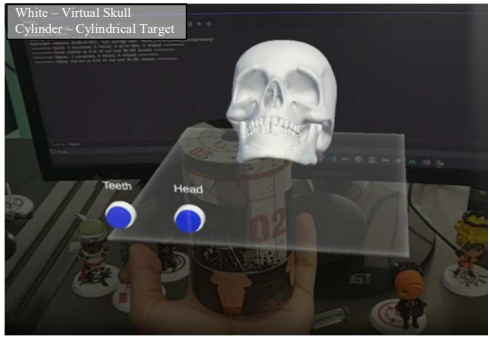

*Figure 5 Tracking the skull from the image target (top) and cylindrical target (bottom)*

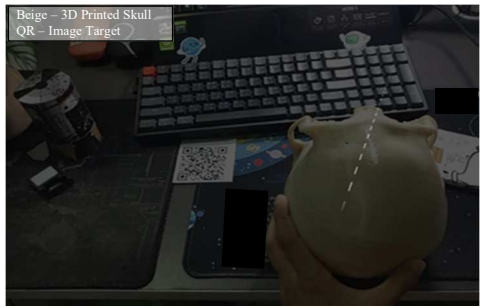

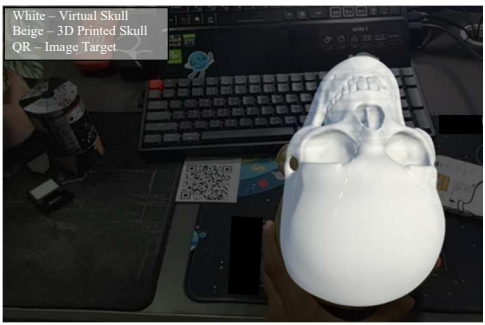

*Figure 6 Successful projection of the virtual skull (bottom) onto the 3D-printed skull (top)*

Following the confirmation of the favorable outcomes derived from the marker-based registration process, wherein the virtual skull was successfully superimposed onto the 3D-printed skull, a subsequent experiment was undertaken.

In this iteration, a skull and its corresponding ventricles underwent segmentation and were then exported as 3D .OBJ files with the help of 3D slicer.

These models were eventually imported into Unity and meticulously arranged, as illustrated in Figure 7. An additional material, characterized by a 40% opacity setting, was applied to the skull model while a green material was applied to the ventricles model to optimize the visibility of the ventricles.

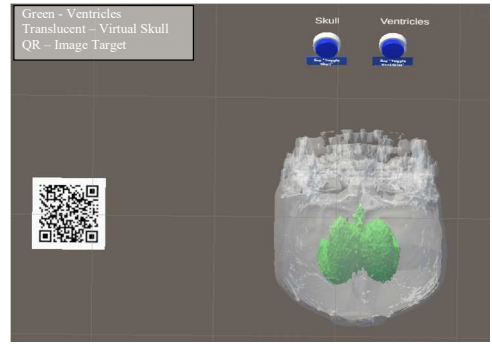

*Figure 7 Successful Unity and Vuforia setup to display the 60% transparent skull and green ventricle models using image target*

Consecutively, both models were linked to buttons and correspondingly associated with voice commands, namely, "Toggle Skull" and "Toggle Ventricles". The placement of these models was precisely executed at a distance of 15 cm to the right of the image target, with the 3D-printed skull situated equidistantly to the right of the marker, as seen in Figure 8.

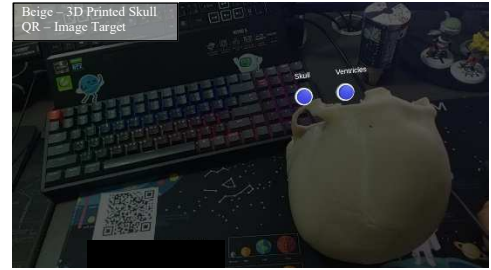

*Figure 8 Positioning the 3D- printed skull 15 cm to the right of the image target*

After a few seconds, the translucent skull and the ventricle highlighted in green were superimposed onto the 3D-printed skull (Figure 9).

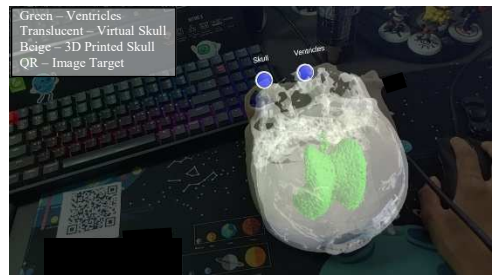

*Figure 9 Successful superimposition of the skull and ventricles onto the 3D-printed skull*

Notably, these models, comprising the skull and ventricle, were made so they could be toggled on and off through voice commands. This feature would allow the surgeons to selectively determine the elements they wish to visualize, as seen in Figure 10.

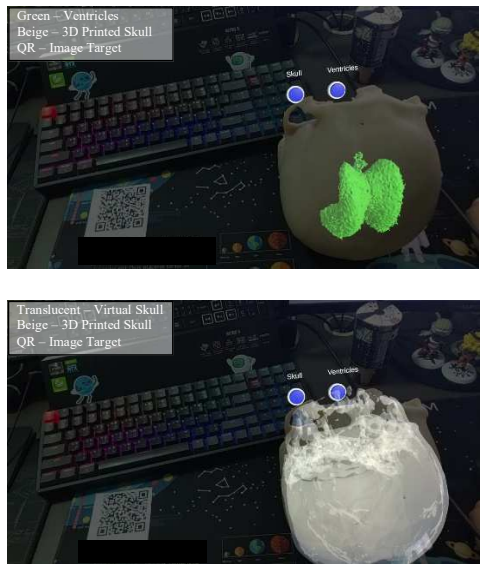

*Figure 10 Toggling off the skull (top) and ventricles (bottom) using voice commands*

Moreover, considering that surgeons predominantly conduct procedures from the perspective of the ventricle targeted for drilling, it became imperative to verify the alignment accuracy of the superimposed skull and ventricle models on the 3D-printed skull when observed from various angles (refer to Figure 11).

Upon inspection of the skulls, it became apparent that the system effectively aligned the virtual models with the 3D-printed counterparts. This alignment, in theory, provides valuable assistance to surgeons in visualizing the precise location of ventricles within the patient's cranial anatomy.

The overall time required, excluding model building, was found to be less than five minutes. Notably, the setup and registration steps took less than a minute each, demonstrating significant efficiency compared to other procedures. The distance between the tracker and the actual skull was consistently within a centimeter margin of error, minimizing fluctuations. While these initial tests validated the potential of employing AR technologies in the proposed solution for neurosurgeries, a more comprehensive experiment was warranted to further substantiate its applicability. Designing meticulous experiments involving surgeons performing VP shunts on 3D-printed skulls derived from patient DICOM files could provide valuable insights. These operations would leverage AR glasses and the proposed flow, with results systematically compared against those generated by VPSEAR, serving as a benchmark for success.

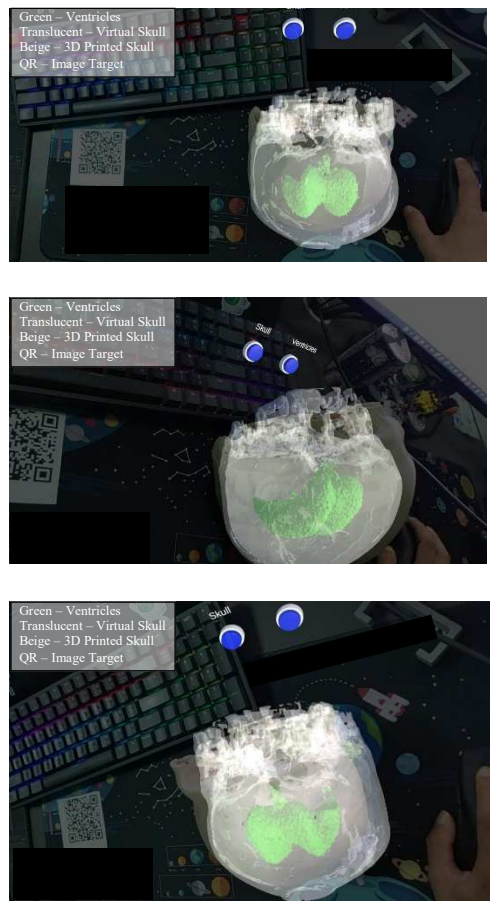

*Figure 11 Examination of the superimposition from top (top), left (middle), and right (bottom)*

Despite the successful results, certain limitations were evident during the experimentation. Dated hardware, particularly the HoloLens 2, exhibited signs of bottlenecking, impacting specific experimental runs. The insufficiency of marker lighting, especially notable in cylindrical target registration, caused delays in virtual model display. While these challenges are inherent to engineering tasks, ongoing advancements in hardware and software are expected to overcome these limitations. Additionally, at present, the process involves the manual generation of new models using 3D slicer, followed by their importation into Unity. Subsequently, the updated build files must be uploaded to HoloLens 2. While the current procedure is not overly time-consuming, there is potential for automation. Implementing an easily navigable user interface would enable users to seamlessly upload CT slides, triggering an automated sequence that encompasses model generation and importing steps. This enhancement would result in users receiving automatic updates of the new models within the AR application.

IV. CONCLUSION

The current landscape of neurosurgical assistance globally faces challenges due to the expense, time consumption, and need for bulky machinery. These constraints limit the adoption of best practices in hospitals, increasing patient risks. To counter this, the proposed solution introduces a new pre-surgery flow and an AR application. This innovation aims to simplify and cut costs while delivering comparable or better results, primarily requiring an investment in AR devices, significantly less than alternative machinery expenses.

This paper has delved into the exploration of integrating AR into VP shunt planning, revealing promising results. From studies in this work, we discovered that the use of AR glasses for the pre-surgery phase was plausible. At this stage, clear visualization of the ventricle and skull anatomy of the patient via the AR model has been shown to be substantially helpful. The heightened spatial understanding of the relationships within the patient's anatomy facilitates surgeons in making informed real-time decisions. This capability enables surgeons to dynamically adjust and refine their shunt planning based on immediate visual feedback.

Our future plan is to map the entry point of Keen or VPSEAR's entry area to the AR model and provide a drilling trajectory to prove or disprove successful intercept to the ventricle.